\documentclass[10pt,twocolumn,letterpaper]{article}

\usepackage[margin=1in]{geometry}
\usepackage{fontspec}
\usepackage{microtype}
\usepackage{xcolor}

\usepackage{amsmath,amssymb,mathtools}
\DeclareMathOperator{\softplus}{softplus}
\DeclareMathOperator{\clip}{clip}

\usepackage{graphicx}
\usepackage{booktabs}
\usepackage{multirow}
\usepackage{siunitx}
\usepackage{float}
\usepackage{dblfloatfix}
\usepackage{array}
\usepackage{tabularx}
\usepackage{flushend}
\usepackage[font=small,labelfont=bf]{caption}
\usepackage{subcaption}

\usepackage{enumitem}
\setlist{nosep,leftmargin=1.2em}

\usepackage{listings}
\lstdefinelanguage{json}{
morestring=[b]", morecomment=[l]{//}, sensitive=true,
showstringspaces=false, breaklines=true,
morekeywords={true,false,null}
}
\lstdefinestyle{papercode}{
basicstyle=\ttfamily\footnotesize,
numbers=left, numberstyle=\tiny, numbersep=6pt,
showstringspaces=false, breaklines=true, breakatwhitespace=true, keepspaces=true,
frame=single, framerule=0.4pt, tabsize=2, columns=fullflexible,
aboveskip=6pt, belowskip=6pt
}
\usepackage[ruled,vlined,linesnumbered]{algorithm2e}
\SetAlFnt{\footnotesize}
\SetCommentSty{textit}
\SetKwComment{tcp}{// }{}
\SetKwInput{KwIn}{Input}
\SetKwInput{KwOut}{Output}

\usepackage{cite}
\newcommand{\supercite}[1]{\textsuperscript{\cite{#1}}}
\let\citet\cite \let\citep\cite

\usepackage{acro}
\acsetup{first-style=short-long, single=true}
\DeclareAcronym{dpvoqat}{
short = {DPVO\mbox{-}QAT++},
long  = {DPVO-QAT++: Heterogeneous QAT and CUDA Kernel Fusion for High-Performance Deep Patch Visual Odometry}
}
\newcommand{\sys}{\ac{dpvoqat}}

\usepackage{hyperref}
\usepackage{bookmark}
\usepackage[nameinlink,capitalise]{cleveref}

\hypersetup{
  breaklinks=true,            
  colorlinks=true,
  linkcolor=blue!60!black,
  citecolor=blue!60!black,
  urlcolor=blue!60!black,
  pdfauthor={Cheng Liao},
  pdftitle={DPVO-QAT++: Heterogeneous QAT and CUDA Kernel Fusion for High-Performance Deep Patch Visual Odometry}
}

\usepackage{xurl}             
\makeatletter
\g@addto@macro\UrlBreaks{\do\/\do\-}
\makeatother

\newcolumntype{Y}{>{\centering\arraybackslash}X}

\usepackage{placeins}

\title{\textbf{DPVO\mbox{-}QAT++}: Heterogeneous QAT and CUDA Kernel Fusion for High\mbox{-}Performance Deep Patch Visual Odometry}
\author{Cheng Liao}
\date{\today}

\usepackage{indentfirst}
\begin{document}
\maketitle

\begin{abstract}
Deep learning-based Visual SLAM (vSLAM) systems exhibit exceptional geometric reasoning capabilities, yet their prohibitive computational
overhead severely restricts deployment on resource-constrained autonomous platforms. This paper presents a hierarchical quantization
optimization framework, \sys. Through the synergistic integration of learnable scale parameterization, a heterogeneous precision design
for the Visual Odometry (VO) front-end and back-end (front-end floating-point fake quantization with FP16/FP32; back-end full precision),
and GPU-native kernel fusion for fake quantization (custom CUDA kernels), our framework significantly reduces memory footprint and
increases processing speed while preserving the trajectory accuracy of the original model. On the TartanAir dataset, our framework
achieves an average FPS increase of 52.1\%, a 29.1\% reduction in median latency, and a 64.9
while maintaining trajectory accuracy (ATE) comparable to the original DPVO model across 32 validation sequences. On the EuRoC dataset,
it realizes an average FPS increase of 30.1\%, a 23.1\% reduction in median latency, and a 37.7\% reduction in peak GPU memory reservation,
maintaining comparable trajectory accuracy (ATE) across 11 validation sequences. Experimental results demonstrate that \sys effectively
bridges the gap between high-precision deep VO and the efficiency requirements for practical deployment, offering a viable engineering
paradigm for the application of this technology on real-world embedded platforms.
\end{abstract}

\section{Introduction}

Visual Simultaneous Localization and Mapping (vSLAM) is a key technology for enabling autonomous robots to perceive and understand their own state and surrounding environment, laying the foundation for devices such as autonomous vehicles, unmanned aerial vehicles, and augmented reality to navigate autonomously in unknown environments.\supercite{DurrantWhyteBailey2006SLAMPartI, Cadena2016SLAMRobustPerception} Classical methods, such as ORB-SLAM\supercite{MurArtal2015ORBSLAM} and LSD-SLAM,\supercite{Engel2014LSDSLAM} rely on hand-crafted features and geometric optimization models based on traditional mathematical principles. Although these methods can run efficiently in real-time on standard CPUs, their robustness is often compromised in challenging scenarios such as low-texture areas, drastic illumination changes, or in the presence of numerous dynamic objects.\supercite{MurArtal2017ORBSLAM2,Almalioglu2022VSLAMSurvey} In recent years, the fusion of deep learning with traditional vSLAM has significantly improved system accuracy and robustness.\supercite{Zhang2024HiSLAM} A representative model, DPVO (Deep Patch Visual Odometry),\supercite{TeedDeng2023DPVO} has achieved state-of-the-art (SOTA) performance on multiple benchmarks with its novel dense patch-tracking network and end-to-end differentiable Bundle Adjustment (BA) back-end. However, its exceptional performance also comes with substantial computational and memory overhead. This creates a significant "deployment gap" for resource-constrained embedded platforms, where theoretically powerful models are difficult to run efficiently in real-time in practical applications.\supercite{WeichaoShi2025SAE}

To bridge this deployment gap, we argue that a co-design methodology that balances algorithmic optimization and hardware acceleration is essential. To this end, we propose \textsc{DPVO-QAT++}, a hierarchical deployment-optimization framework for the DPVO system. The core contributions of this framework include: (1) designing a geometry-aware "scale-only" Quantization-Aware Training (QAT) strategy that learns transferable quantization scales while keeping the model weights in floating-point format, thereby maximally preserving the geometric representation capabilities of the pre-trained model;\supercite{Jacob2018IntegerOnlyInference} (2) proposing a heterogeneous precision architecture that applies pseudo-quantization (implemented in floating-point) only to the compute-intensive network front-end, while strictly maintaining the numerically sensitive geometric optimization back-end in full precision, thus striking a balance between efficiency and accuracy; and (3) developing custom CUDA kernels that fuse pseudo-quantization operators to eliminate framework scheduling overhead, thereby accelerating the floating-point simulation of low-precision behavior.\supercite{Boehm2014SystemMLDEB}

Through the synergistic optimization of its algorithm and system architecture, this work transforms DPVO into a practical system that balances both high accuracy and high efficiency. Experimental results show that DPVO-QAT++ significantly improves processing frame rate, reduces real-time latency, and substantially decreases memory footprint, all while maintaining trajectory accuracy nearly identical to the original model. This not only validates the effectiveness of our proposed methodology but also provides a generalizable and reproducible optimization paradigm for deploying complex perception algorithms on real-world robotic platforms.

\section{Related Work}
\subsection{The Evolution of Visual SLAM: From Geometry to Learning}
The evolution of vSLAM is synchronized with the core technological transformations in computer vision and robotics, with its theoretical foundations tracing back to the probabilistic frameworks of the 1980s.\supercite{MurArtal2017ORBSLAM2} In the early 21st century, driven by increased computational power, vSLAM entered the "Geometric Era," which diverged into two classical schools: feature-based methods, represented by MonoSLAM\supercite{Davison2007MonoSLAM} and ORB-SLAM\supercite{MurArtal2015ORBSLAM}, which estimate pose by extracting and matching sparse points followed by back-end optimization;\supercite{MurArtal2017ORBSLAM2} its sophisticated multi-threaded architecture (tracking, local mapping, loop closure detection) enabled real-time performance on CPUs, establishing it as a key benchmark for subsequent research.\supercite{Campos2021ORBSLAM3} The other school consists of direct methods, represented by LSD-SLAM,\supercite{Engel2014LSDSLAM} which construct semi-dense maps by minimizing photometric error.\supercite{Engel2015StereoLSDSLAM} Although these classical methods are exceptionally efficient, their reliance on hand-crafted modules limits their robustness in complex scenarios.\supercite{Almalioglu2022VSLAMSurvey}

In recent years, the wave of deep learning has propelled vSLAM into the "Learning Era."\supercite{Li2018OngoingEvolutionVSLAM} Neural networks are now widely used to replace traditional modules for tasks like feature extraction,\supercite{Sarlin2020SuperGlue} depth estimation,\supercite{Ranftl2022TrobustDepth} or optical flow prediction,\supercite{TeedDeng2020RAFT} culminating in end-to-end deep VO/SLAM systems that treat pose estimation as a regression problem directly solvable by a neural network.\supercite{Zhang2024HiSLAM} DPVO is a landmark achievement in this direction, refreshing state-of-the-art accuracy on multiple benchmarks with its sparse patch tracking technique tightly coupled to a differentiable BA back-end.\supercite{TeedDeng2023DPVO,Teed2024DPVSLAM} However, its success also exposes a common challenge of deep methods: immense computational and memory overhead,\supercite{Boehm2014SystemMLDEB} which is the primary conflict this work aims to resolve.

\begin{figure*}[!t]
  \centering
  \includegraphics[width=\textwidth]{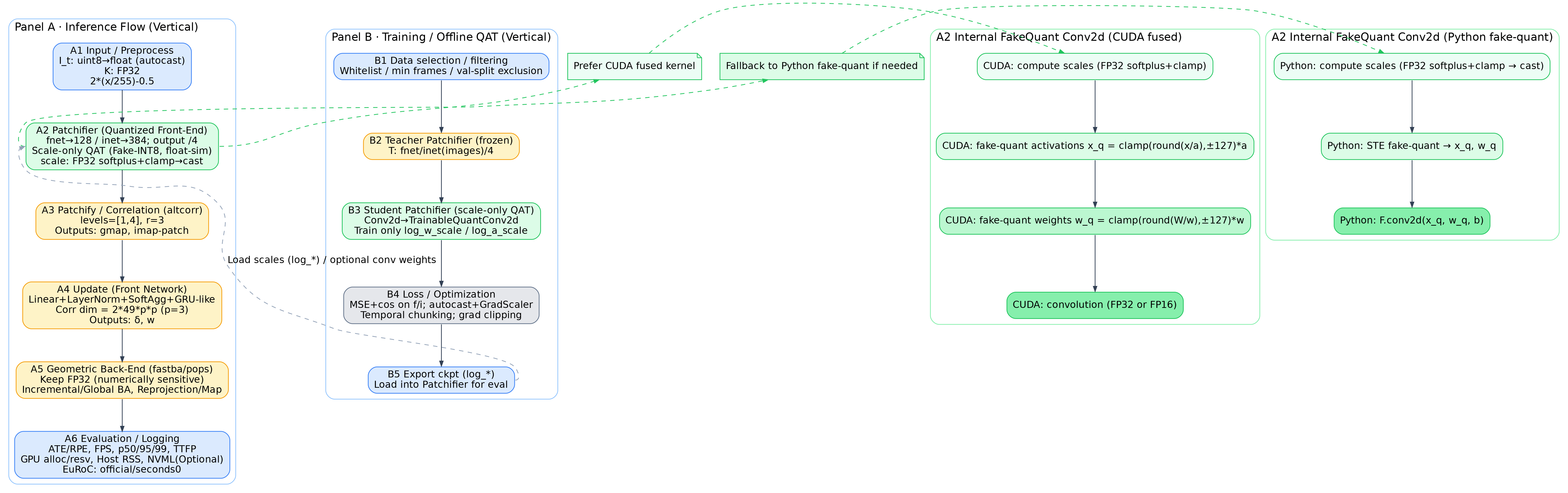}
  \caption{
    Layered deployment and optimization framework for DPVO.
    Panel A shows the main inference flow (A1$\rightarrow$A6): from input preprocessing to the quantized front-end Patchifier (floating-point simulation of scale-learning QAT), followed by feature extraction and correlation, then front-end update, then the FP32 geometric back end (BA/reprojection/map), and finally evaluation and logging. Under A2, two implementations of the internal fake-quantized convolution (\texttt{Conv2d}) are shown: left, CUDA-fused (compute scales $\rightarrow$ fake-quantize activations $\rightarrow$ fake-quantize weights $\rightarrow$ convolution) and right, per-operator Python (compute scales $\rightarrow$ STE fake-quantization $\rightarrow$ \texttt{F.conv2d}). Panel B depicts the offline QAT training pipeline, where only the scale parameters for weights and activations (\texttt{log\_w\_scale}/\texttt{log\_a\_scale}) are learned and then injected into the Patchifier at evaluation time. Color coding: green denotes the front-end QAT/fake-quant; ochre denotes the native DPVO CUDA back end; blue denotes I/O and evaluation; gray denotes loss/optimization.
  }
  \label{fig:framework_diagram}
\end{figure*}

\subsection{Neural Network Quantization and Acceleration for Embedded Deployment}
Neural network quantization is the dominant paradigm for efficient deep learning deployment. Among the approaches, Post-Training Quantization (PTQ) is simple but often suffers from severe accuracy degradation due to error accumulation,\supercite{WeichaoShi2025SAE} whereas Quantization-Aware Training (QAT) achieves a superior accuracy-efficiency trade-off by simulating quantization during training.\supercite{Jacob2018IntegerOnlyInference} Although advanced paradigms such as mixed-precision,\supercite{Davison2007MonoSLAM} learnable,\supercite{Campos2021ORBSLAM3} and hardware-aware\supercite{Engel2015StereoLSDSLAM} quantization have shown significant gains, existing methods are mostly designed for classification/detection tasks, with limited validation in geometric reasoning applications. Visual odometry, with its stringent requirements for temporal consistency, sub-pixel accuracy, and geometric constraints, demands specialized quantization strategies.

At the system level, hardware acceleration is indispensable for deploying models efficiently on edge devices, in addition to algorithmic optimization.\supercite{WeichaoShi2025SAE} Although high-level libraries like cuDNN and TensorRT provide highly optimized GPU implementations for standard network layers,\supercite{ZhouYang2022TensorRT} for composite operations like pseudo-quantization, reliance on the per-operator scheduling of high-level frameworks introduces significant overhead, potentially leading to performance worse than the original floating-point model.\supercite{Boehm2014SystemMLDEB,Zhuang2024MonoNN} To address this, we write custom CUDA kernels to fuse the multiple steps of pseudo-quantization into a single kernel execution, thereby eradicating the bottlenecks of scheduling and memory round-trips.

Specifically, we wrote custom CUDA kernels with fused operators: steps in the critical path of pseudo-quantization—such as tensor division, clipping, rounding, and de-quantization—are fused into a single kernel, merging multiple kernel launches and memory accesses into one. Concurrently, we employ a linear traversal to optimize memory access, enhancing bandwidth utilization. The forward pass consists of three kernel stages—scale computation, activation pseudo-quantization, and weight quantization—which, compared to per-operator scheduling, substantially reduces framework scheduling overhead and global memory round-trips.

Current vSLAM optimization research follows two seldom-intersecting paths: one applies quantization and relies on high-level frameworks for acceleration, while the other pursues ultimate performance by writing custom CUDA kernels for full-precision models.\supercite{Teed2024DPVSLAM} This separation highlights a critical research gap: the deep co-design of quantization strategies with low-level, hardware-specific kernel optimizations.\supercite{Cui2024FusionSurvey} This work aims to bridge the engineering pipeline from algorithm to system through hardware-software co-design, providing a viable optimization paradigm for building vSLAM systems that possess both high accuracy and high efficiency.

\section{DPVO-QAT++: A Heterogeneous Quantization and Acceleration Framework}

To enhance the deployment efficiency of DPVO on resource-constrained edge platforms while preserving its high accuracy, we designed the DPVO-QAT++ framework. This framework completes Quantization-Aware Training (QAT) parameter learning and inference validation by simulating INT8 noise via fake quantization on PC hardware. It represents a comprehensive solution integrating optimizations at the algorithmic, architectural, and system levels(see Fig.~\ref{fig:framework_diagram}).


\subsection{Framework Overview}

The overall workflow of DPVO-QAT++ is divided into two stages: offline training and online inference.

Offline Quantization-Aware Training: In this stage, we employ a teacher-student distillation method. The full-precision DPVO front-end serves as the teacher network, while a duplicated student network acts as the candidate model to be quantized. During the training phase, the student's convolutional kernels and normalization parameters are frozen; only the scales for symmetric S8 fake quantization of activations/weights (stored in log space) are optimized. Through a multi-objective constraint using both Euclidean distance and cosine similarity, the student's output is made to closely follow the teacher's predictions in the feature space. This "scale-only" strategy explicitly models INT8 saturation and clipping errors on one hand, and on the other, avoids corrupting the original geometric representations.\supercite{Peng2024QVIO} Consequently, the learned scales can be directly used as a proxy for the subsequent true INT8 mapping, eliminating the need for repeated tuning or retraining on the target hardware.

Online Accelerated Inference: During the inference stage, an input RGB sequence first enters the Patchifier front-end (fnet/inet). We use a set of custom CUDA kernels to perform fused pseudo-quantization operations separately for scale computation, activations (per-tensor), and weights (per-channel): using a small number of kernels: one for scales, one for activation quantization, and one for weight quantization (each kernel fuses division, clipping, rounding, and de-quantization for that tensor type), significantly reducing per-operator scheduling and intermediate memory access overhead. The convolution computation remains on the floating-point (FP16/FP32) pipeline, with simulated pseudo-quantization error superimposed. The quantized feature and descriptor maps are then directly fed into the full-precision (FP32) back-end geometric optimizer to solve the non-linear constraints, forming a heterogeneous precision topology of a "quantized (floating-point pseudo-quantization) front-end + full-precision back-end".\supercite{Peng2024QVIO} The overall data pathway is: RGB Image → Fake-Quantized Front-End (CUDA fused operators, FP16/FP32 + INT8 error simulation) → Features/Descriptors → Full-Precision Back-End → Pose Estimation.

The core idea of this workflow is heterogeneous precision processing: applying simulated quantization acceleration to the compute-intensive front-end, which has some tolerance to noise, while maintaining the back-end optimizer, which demands high numerical precision, in full precision. This layered architecture not only allows for the accurate assessment of quantization noise on SLAM convergence on PC hardware but also ensures that the learned scale parameters can be seamlessly transferred to subsequent embedded INT8 inference. This enables us to complete parameter freezing and error calibration before actual hardware deployment, avoiding costly retraining processes on resource-constrained platforms and thus creating a closed loop of algorithm validation, system porting, and online debugging.

\subsection{Geometry-Preserving Quantization-Aware Training}
\label{sec:qat-geometry}

\textbf{Motivation and goal.}
In vSLAM, small errors introduced by the front-end feature extraction can be amplified by the back-end iterative optimization, potentially causing catastrophic trajectory drift.\supercite{Peng2024QVIO} Our primary objective in QAT is to \emph{preserve the model's geometric reasoning}.

\textbf{Rationale for a heterogeneous-precision architecture.}
DPVO naturally fits a divide-and-conquer, heterogeneous-precision design. The front end consists of two ResNet-like encoders that extract dense features and dominate overall computation, thus becoming the main target for quantization; the back end performs GatedResidual-based recurrent updates (GRU-like) and differentiable bundle adjustment, which are lighter in FLOPs yet highly sensitive to numerical perturbations. Hence, we \emph{quantize the front end and keep the back end in full precision} to balance efficiency and robustness.

\textbf{Scale-Only QAT.}
Standard QAT updates both network weights and quantization parameters, which can move the solution away from a strong pretrained optimum. To maximally retain the geometric knowledge encoded in DPVO's pretrained weights, we \emph{freeze all convolutional weights} and learn only the quantization scales $s$. We use per-channel weight quantization and per-tensor activation quantization. The log-domain parameters are:
\begin{gather}
  \log s_w \in \mathbb{R}^{C_{\text{out}}}, \label{eq:logsw}\\
  \log s_a \in \mathbb{R}. \label{eq:logsa}
\end{gather}
The actual forward scales are
\begin{align}
  s_w &= \clip\!\big(\softplus(\log s_w)+\epsilon,\ s_{\min},\,s_{\max}\big), \label{eq:sw}\\
  s_a &= \clip\!\big(\softplus(\log s_a)+\epsilon,\ s_{\min},\,s_{\max}\big). \label{eq:sa}
\end{align}

\textbf{Fake quantization with STE.}
We simulate quantization in the forward pass and use a straight-through estimator (STE) for backpropagation:\supercite{Esser2020LSQ}
\begin{align}
  \mathrm{FQ}(x,s)
  &= s \cdot \mathrm{round}\!\Big(
     \clip\big(x/s,\,-q_{\max},\,q_{\max}\big)
     \Big). \label{eq:fq}
\end{align}
For signed INT8 ($b{=}8$),
\begin{equation}
  q_{\max}=2^{\,b-1}-1=127. \label{eq:qmax}
\end{equation}

\textbf{Teacher--student distillation and joint loss.}
We adopt the original DPVO as the teacher. The quantized student learns proper scales via a joint loss enforcing \emph{numerical proximity} (MSE) and \emph{directional consistency} (cosine similarity), which is critical for patch matching\supercite{TungMori2019SPKD}:
\begin{equation}
\begin{aligned}
  \mathcal{L} \;=\;&\ \mathrm{MSE}(F_s,F_t)
  \;+\; \mathrm{MSE}(\widehat{I}_s,\widehat{I}_t) \\
  &+\; \lambda_{\mathrm{cos}}\!\big(1-\mathrm{CosSim}(F_s,F_t)\big) \\
  &+\; \lambda_{\mathrm{cos}}\!\big(1-\mathrm{CosSim}(\widehat{I}_s,\widehat{I}_t)\big).
\end{aligned}
\label{eq:loss}
\end{equation}

\textbf{Training details and implementation.}
During training, only $\log s_w$ and $\log s_a$ in the front-end \texttt{TrainableQuantConv2d} layers are updated. All normalization layers are forced to \texttt{eval} at the beginning, freezing their statistics and affine parameters. The scale-only strategy is implemented as follows: the optimizer explicitly collects scale parameters via \texttt{collect\_scale\_params} and applies Adam to them only, leaving all other weights frozen to preserve pretrained geometric features.




\subsection{GPU-Native Acceleration for Front-End Inference}
\label{sec:gpu-native-accel}

While QAT implementations in standard deep learning frameworks are functionally complete, their runtime overhead can be substantial—at times even making the quantized model \emph{slower} than the original floating-point baseline.\supercite{Zhang2019CUDAKernelOverheads,Lustig2013HPCA} A key reason is that fake-quantization is decomposed into multiple \emph{separate} GPU kernel launches, which introduces nontrivial scheduling and memory-traffic overhead.

\textbf{Bottleneck analysis of fake quantization.}
In PyTorch-like frameworks, one fake-quantization step (cf.\ Eq.~\eqref{eq:fq}) is typically realized as four sub-ops: division, clipping, rounding, and multiplication. Each sub-op triggers a kernel launch and global-memory reads/writes. Modern GPUs are highly inefficient for such \emph{pico-kernels}: arithmetic intensity is tiny, and most time is lost in CPU$\rightarrow$GPU dispatch and inter-kernel waits.\supercite{Wahib2014KernelFusion} Empirically, a straightforward QAT-DPVO (without CUDA fusion) \emph{regresses} in throughput versus the original model, directly confirming this bottleneck.

\textbf{Python--CUDA integration.}
\begin{sloppypar}
We bridge Python-side evaluation and low-level CUDA kernels via a PyTorch C++ extension.\supercite{Paszke2019PyTorch} In the evaluation script, model initialization performs a ``QAT injection'': it recursively traverses
the DPVO front end (the \emph{patchifier}) and wraps each \texttt{nn.Conv2d} with
\texttt{EvalOptimized\-TrainableQuant\-Conv2d} (preserving the original \texttt{Conv2d} instance and weights). The wrapper first
invokes the compiled C++/CUDA extension to execute fused fake-quantization on GPU, then calls ATen/cuDNN \texttt{conv2d}.
If the extension is unavailable or throws, it automatically falls back to Python-side fake-quantization plus \texttt{F.conv2d}.
\end{sloppypar}

\textbf{Custom fused CUDA kernels.}
\begin{sloppypar}
To remove the above overhead at its source, we implement fused CUDA kernels along the critical path of fake-quantization.
The design fuses the four steps for activations and for weights \emph{within single kernels} (and adds one kernel to
compute scales via softplus+clamp on GPU), while the convolution itself is executed by ATen/cuDNN \texttt{conv2d}.
Compared with unfused pico-kernels and repeated global-memory round-trips, the pipeline becomes:
\emph{compute scales once} $\rightarrow$ \emph{fake-quantize activations} $\rightarrow$ \emph{fake-quantize weights}
$\rightarrow$ \emph{one high-performance convolution call}, significantly cutting launch and memory-traffic overhead.\supercite{Jang2011MemoryAccessPatterns}
For kernel efficiency, we adopt grid–stride loops and contiguous indexing patterns that are friendly to memory coalescing;
vectorized I/O (e.g., \texttt{float4} loads/stores) is an optional optimization for embedded deployments.
\end{sloppypar}




\textbf{Mixed precision and robustness.}
For numerical stability, the CUDA path performs the required FP16$\leftrightarrow$FP32 conversions;\supercite{Micikevicius2018MixedPrecision} by default,
inputs and weights are promoted to FP32 before entering the extension, fake-quantization executes in C++, and the convolution is
delegated to ATen/cuDNN, so inference follows an FP32 path. For engineering robustness, if the extension is unavailable
or any exception is raised, the system automatically falls back to Python fake-quantization plus \texttt{F.conv2d};
whether the fallback uses FP16 autocast is controlled by higher-level configuration.

\section{Experiments}\label{sec:experiments}

This chapter systematically evaluates \textsc{DPVO--QAT++} on both real and synthetic datasets and provides code- and weight-level evidence for verification. We first describe the experimental setup (\S\ref{subsec:setup}), then present quantitative results on system performance, memory efficiency, and trajectory accuracy with visualizations (\S\ref{subsec:results}), and conclude with ablation studies to validate the contribution of each core component (\S\ref{sec:ablations}).

\subsection{Experimental Setup}\label{subsec:setup}

\subsubsection{Datasets and Benchmarks}\label{subsubsec:datasets}

\paragraph{EuRoC MAV.}
A widely used real-world benchmark captured in indoor environments, providing strictly time-synchronized stereo imagery, IMU measurements, and high-precision ground truth from a motion-capture system.\supercite{Burri2016EuRoC} To follow the official \textsc{DPVO} evaluation protocol, we use the left monocular stream (cam0). This dataset primarily tests whether \textsc{DPVO--QAT++} preserves accuracy and runtime efficiency under deployment-like conditions.

\paragraph{TartanAir.}
A high-fidelity synthetic SLAM environment covering diverse scenes, illumination conditions, and extreme motions.\supercite{Wang2020TartanAir} We use it to examine robustness and generalization under challenging, non-ideal conditions. For plot readability, sequence names are slightly shortened in figures (e.g., \texttt{amusement-Easy-08} $\rightarrow$ \texttt{amusement-E}); full name mappings are provided in the plotting scripts.

\subsubsection{Evaluation Metrics}\label{subsubsec:metrics}

We use three categories of metrics for a comprehensive evaluation:

\paragraph{Accuracy.} Absolute Trajectory Error (ATE) RMSE via the \textit{evo} toolchain, with rigid alignment and scale correction.\supercite{Sturm2012RGBDBenchmark}

\paragraph{Performance.} Average FPS for throughput; latency P50/P99 (ms) for steady-state and tail response; time-to-first-pose (TTFP, ms) from first frame arrival to first pose for initialization efficiency.

\paragraph{Resource Usage.} Peak GPU \texttt{Allocated}/\texttt{Reserved} memory (GB) and host process RSS (GB) to assess memory efficiency and deployment feasibility.

\subsubsection{Hardware and Software Platform}\label{subsubsec:platform}

All experiments are run on the same device: RTX~4060 (8\,GB) + Intel Core Ultra~5--125H + 32\,GB RAM; software stack: Python~3.10 + PyTorch~2.3.1 + CUDA~12.1. During inference we use ``floating-point convolution + INT8 error simulation (Q/DQ)'', with convolutions executed by cuDNN float kernels. The front-end fake-quantization supports two paths: standard PyTorch per-operator fake-quant and our fused CUDA path (main results use the latter; see \S\ref{subsec:throughput-latency}). To ensure determinism and cross-platform reproducibility, we disable options that may introduce non-determinism, including TensorFloat-32 and \texttt{cudnn.benchmark}.

\subsubsection{Baselines}\label{subsubsec:baselines}

\paragraph{Baseline \textsc{DPVO}.} Official implementation (FP16/FP32 mixed precision), no quantization or custom kernels, default (non-\texttt{fast}) inference.\supercite{TeedDeng2023DPVO}

\paragraph{\textsc{DPVO--QAT++} (Ours).} ``Scale-only'' QAT for the front end with fused CUDA fake-quantization kernels for Q/DQ; convolutions remain on cuDNN/ATen; the back-end geometric optimization stays in FP32.

\paragraph{Note.} The primary comparison is Baseline vs.\ Ours. For ablations we additionally consider \textsc{QAT--DPVO} (algorithm-only, per-operator fake-quant) to isolate the contribution of system-level fusion.

\subsubsection{Training Data and Pipeline}\label{subsubsec:train}

Training data curation and filtering are managed by dedicated scripts. The core training adopts a teacher--student distillation paradigm to align the student's \textit{Patchifier} features at the $\times\!1/4$ scale. Convolutions and normalization layers in the student remain frozen; the optimizer updates only the quantization scales: per-channel weight scales \texttt{log\_w\_scale} and per-tensor activation scales \texttt{log\_a\_scale}. The loss is a weighted sum of MSE and cosine similarity. To ensure stability, the forward pass may use \texttt{autocast(FP16)}, but scale parameters are computed in FP32, then passed through \texttt{softplus} and \texttt{clamp} before being applied to fake quantization. We further adopt temporal chunking and \texttt{isfinite} guards for robustness.

For evaluation, each sequence is run five times and we report the median to reduce randomness.\supercite{TeedDeng2023DPVO} A composite summary table is provided in the supplementary material for comparative studies.

\subsection{System-Level Quantitative Results}\label{subsec:results}

\subsubsection{Key Findings (Overview)}\label{subsub:key-findings}
\begin{itemize}[leftmargin=1.2em]
  \item \textbf{Significant performance gains.} Median per-sequence FPS ratio (Ours/Baseline): TartanAir $\approx$ +34.6\%, EuRoC $\approx$ +26.7\%. Median per-sequence tail-latency improvement (P99): $-22.8$\,ms (TartanAir) and $-19.7$\,ms (EuRoC) (see Fig.~\ref{fig:fps_bars} for FPS, Fig.~\ref{fig:latency_bars} for P99).

\begin{figure}[!t]
  \centering
  \includegraphics[width=\columnwidth]{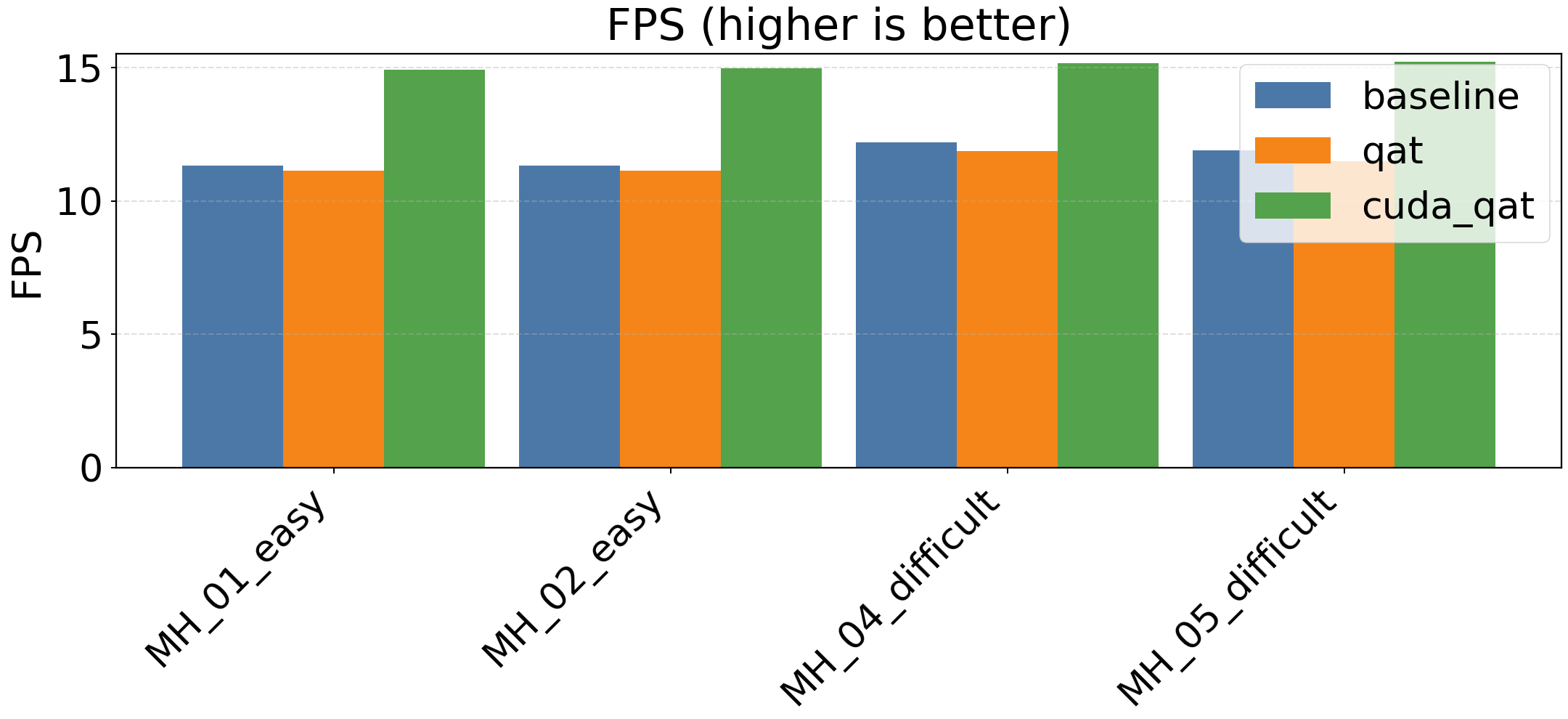}\\[3pt]
  \includegraphics[width=\columnwidth]{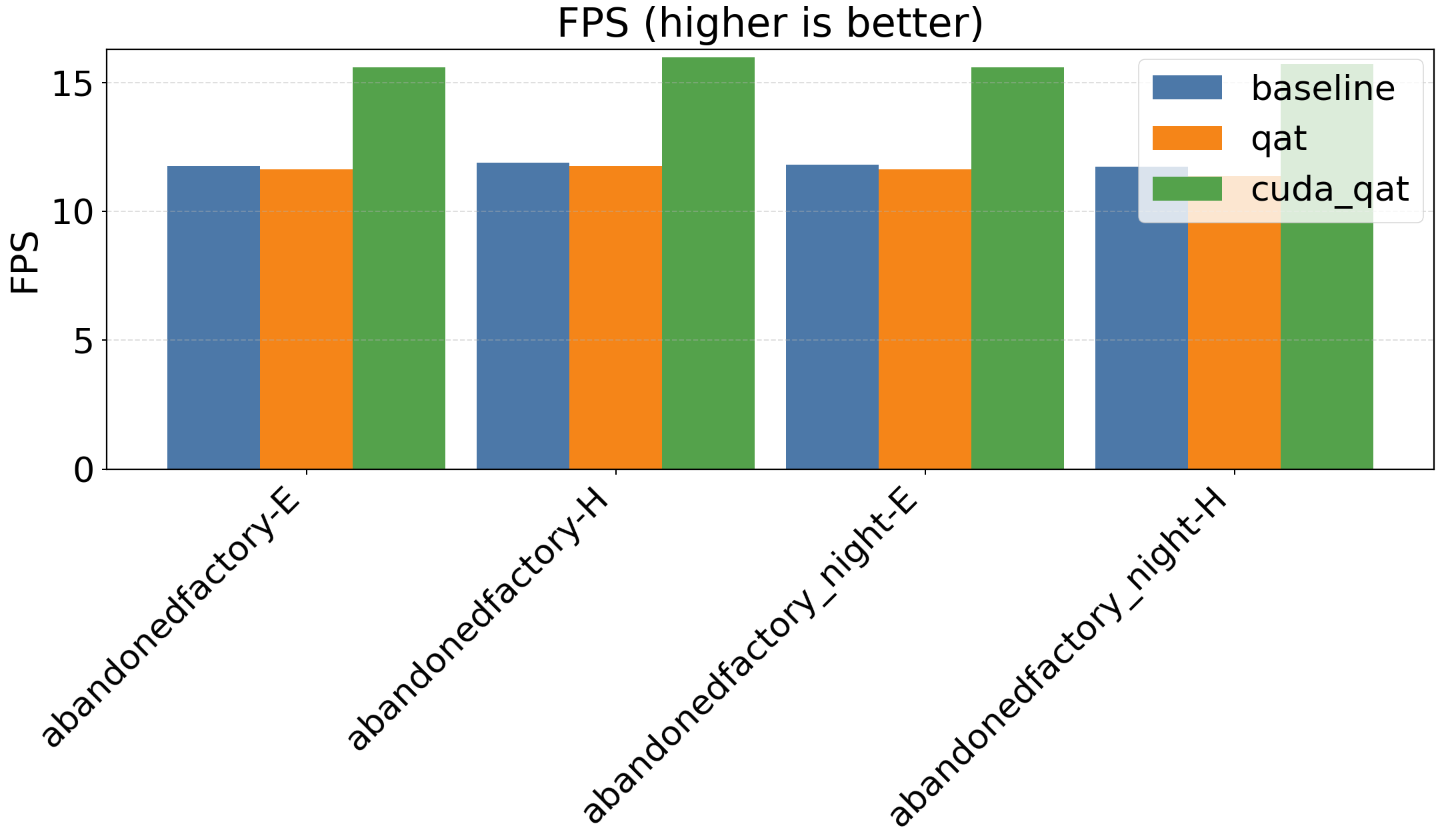}
  \caption{
    Average Frames Per Second (FPS) comparison on EuRoC (top) and TartanAir (bottom). 
    The bar charts compare the steady-state performance of the baseline, our QAT-only approach, and the full \textsc{DPVO--QAT++} framework. Higher FPS is better.
  }
  \label{fig:fps_bars}
\end{figure}

\begin{figure}[!t]
  \centering
  \includegraphics[width=\columnwidth]{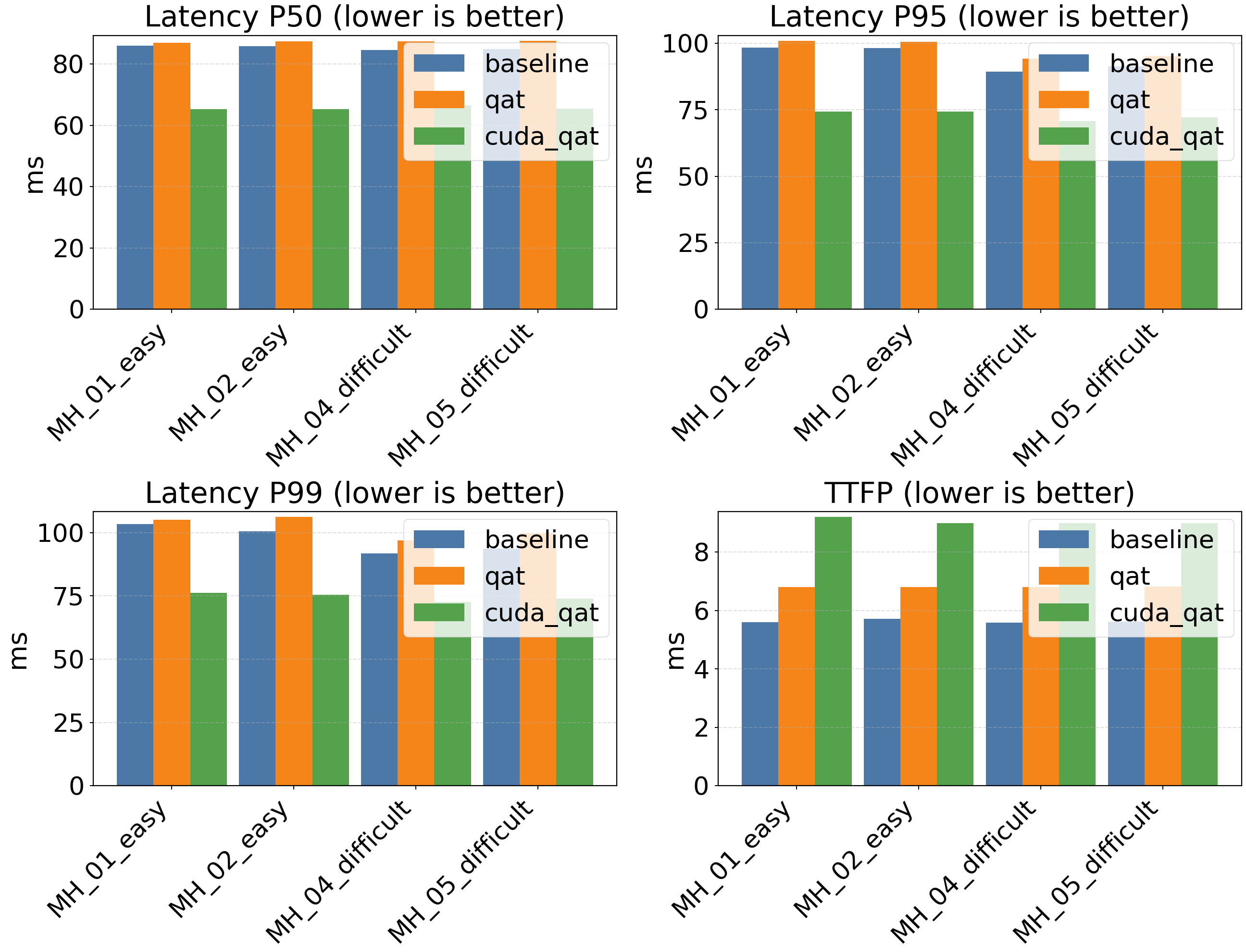}\\[3pt]
  \includegraphics[width=\columnwidth]{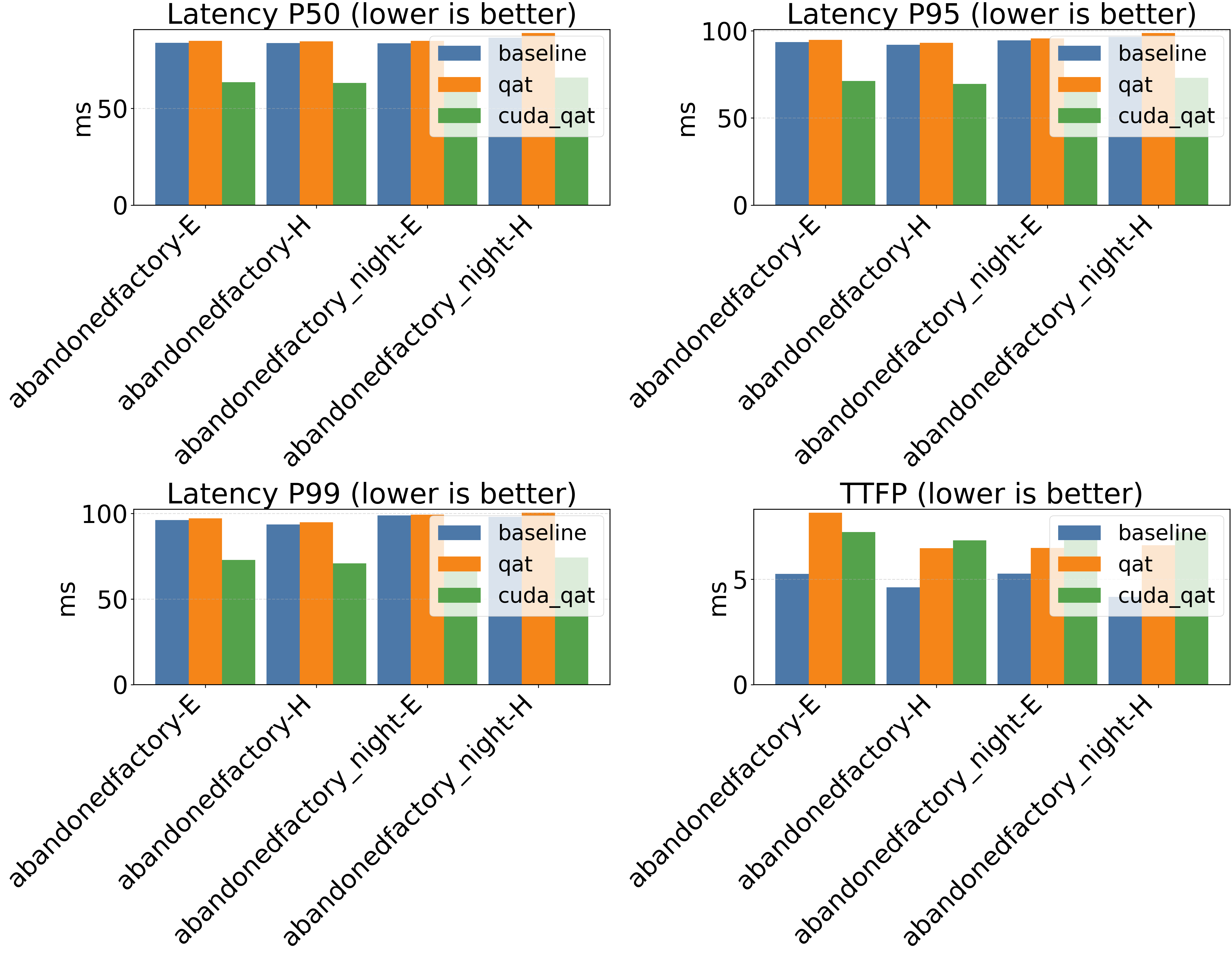}
  \caption{
    Inference latency comparison on EuRoC (top) and TartanAir (bottom). 
    Each chart shows multiple metrics, including steady-state P50, P95, and P99 latency, as well as Time To First Frame (TTFP). Lower latency (in ms) is better.
  }
  \label{fig:latency_bars}
\end{figure}

  \item \textbf{Pure QAT offers limited speedup.} \textsc{QAT--DPVO} with per-operator fake-quant yields flat or slightly lower FPS ($-1\%\sim-3\%$) and slightly higher P99, consistent with expected scheduling and memory-traffic overheads of non-fused paths (see Fig.~\ref{fig:fps_bars} for FPS).
  \item \textbf{Memory efficiency.} Peak \texttt{Reserved} memory drops substantially for \textsc{DPVO--QAT++} (e.g., from 1.94\,GB to 1.02\,GB on a representative TartanAir sequence, $\approx -47\%$), improving deployability on memory-limited devices (see Fig.~\ref{fig:memory_bars} for memory efficiency).

\begin{figure}[!t]
  \centering
  \includegraphics[width=\columnwidth]{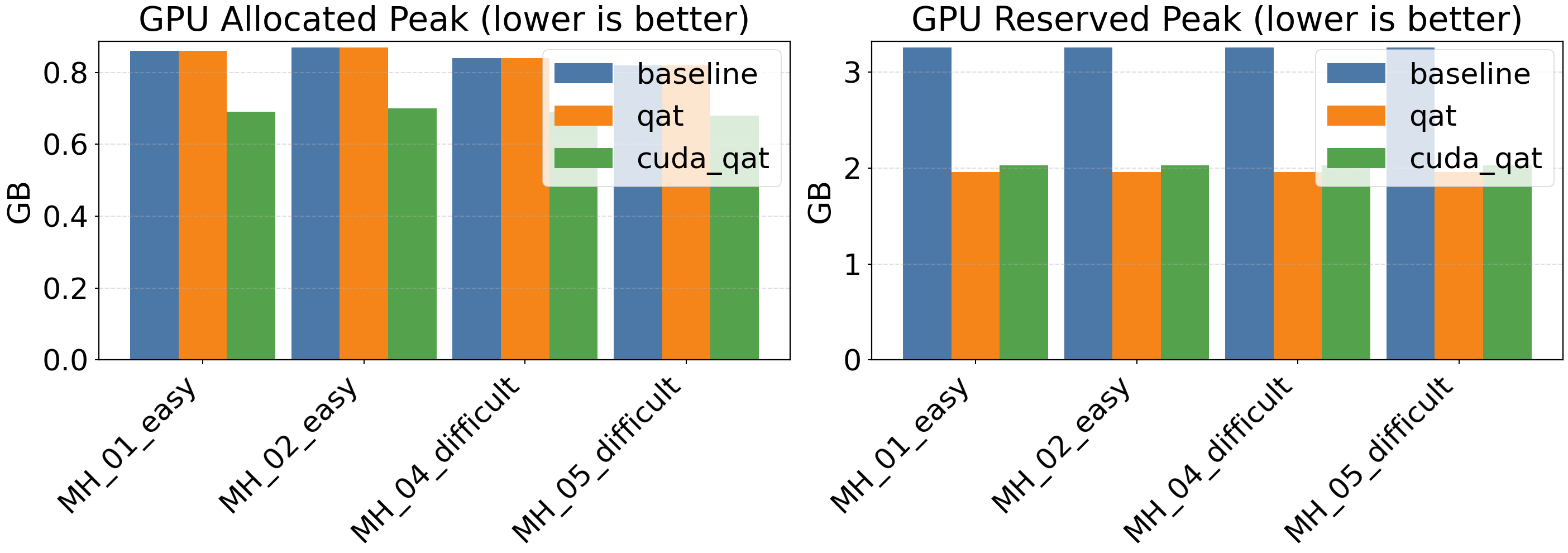}\\[3pt]
  \includegraphics[width=\columnwidth]{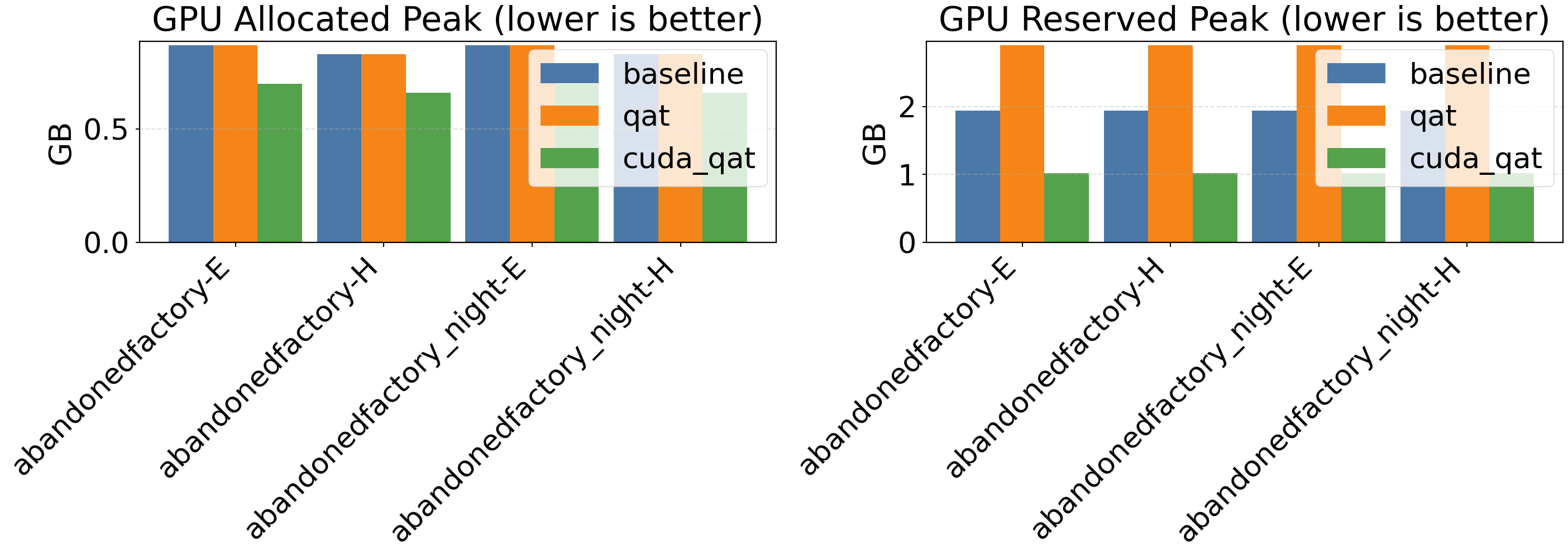}
  \caption{
    Peak GPU memory usage comparison on EuRoC (top) and TartanAir (bottom). 
    The charts show both peak allocated memory (left side of each group) and peak reserved memory (right side of each group). Lower usage (in GB) is better.
  }
  \label{fig:memory_bars}
\end{figure}

  \item \textbf{Accuracy retention.} ATE is on par with the baseline on most sequences and sometimes better, consistent with conservative scale learning and back-end BA absorbing small perturbations (see Fig.~\ref{fig:ate_lines} for ATE trends, Figs.~\ref{fig:combined_overlays} and \ref{fig:combined_overlays} for cross-scene 3D comparisons).
  
\begin{figure}[!t]
  \centering
  \includegraphics[width=\columnwidth]{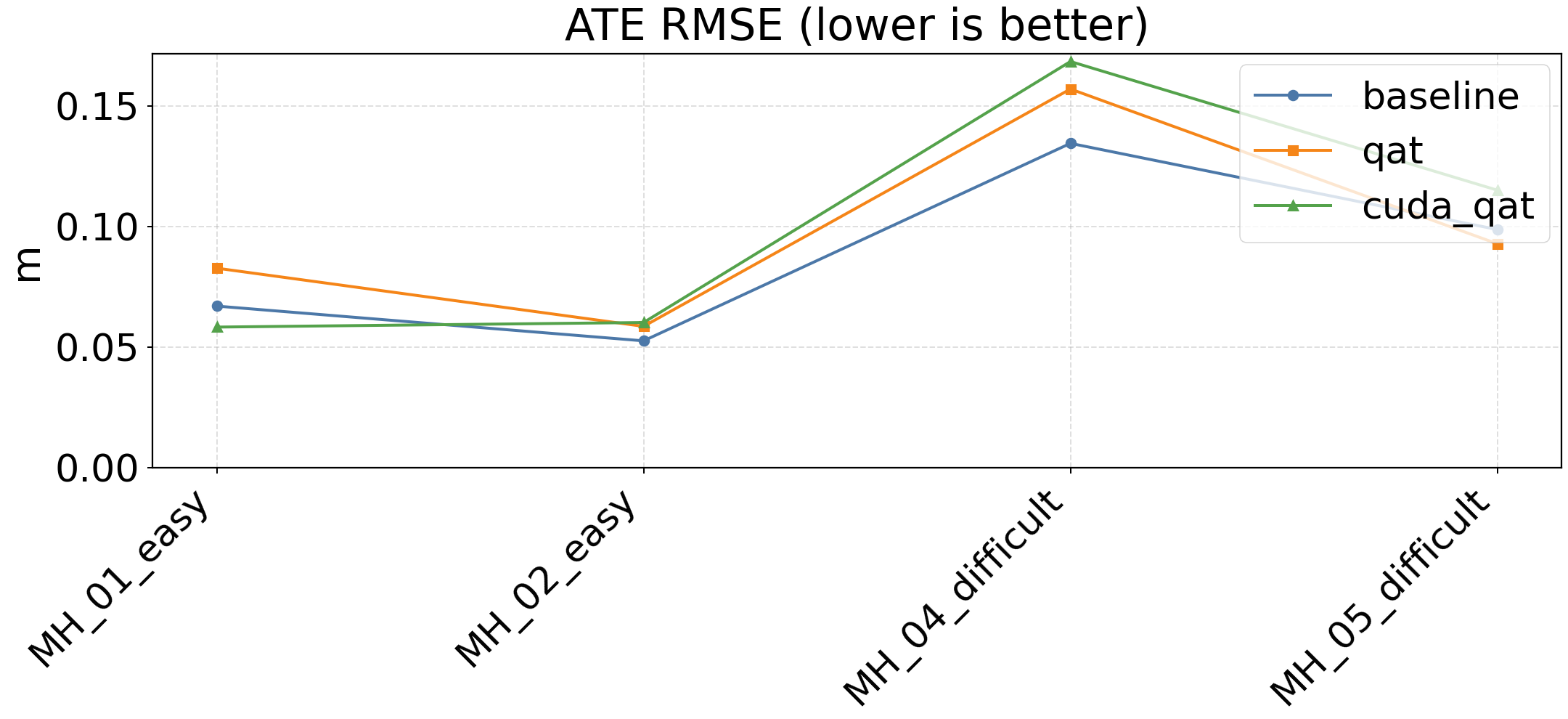}\\[3pt]
  \includegraphics[width=\columnwidth]{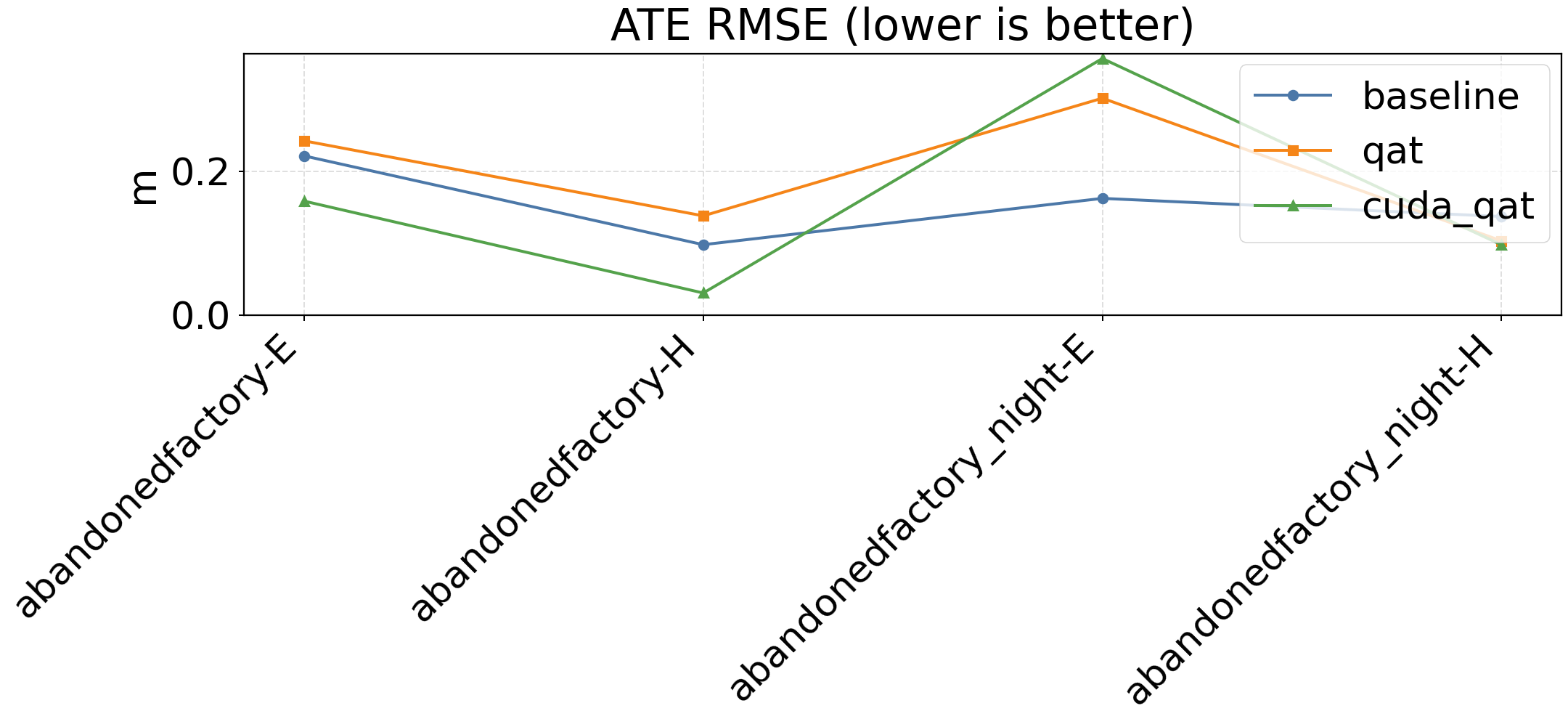}
  \caption{
    Absolute Trajectory Error (ATE RMSE) comparison on EuRoC (top) and TartanAir (bottom). 
    The line charts plot the ATE (in meters) for each sequence, comparing the baseline, QAT-only, and \textsc{DPVO--QAT++} methods. Lower error is better.
  }
  \label{fig:ate_lines}
\end{figure}

\begin{figure}[!t]
  \centering
  
  \includegraphics[width=\columnwidth]{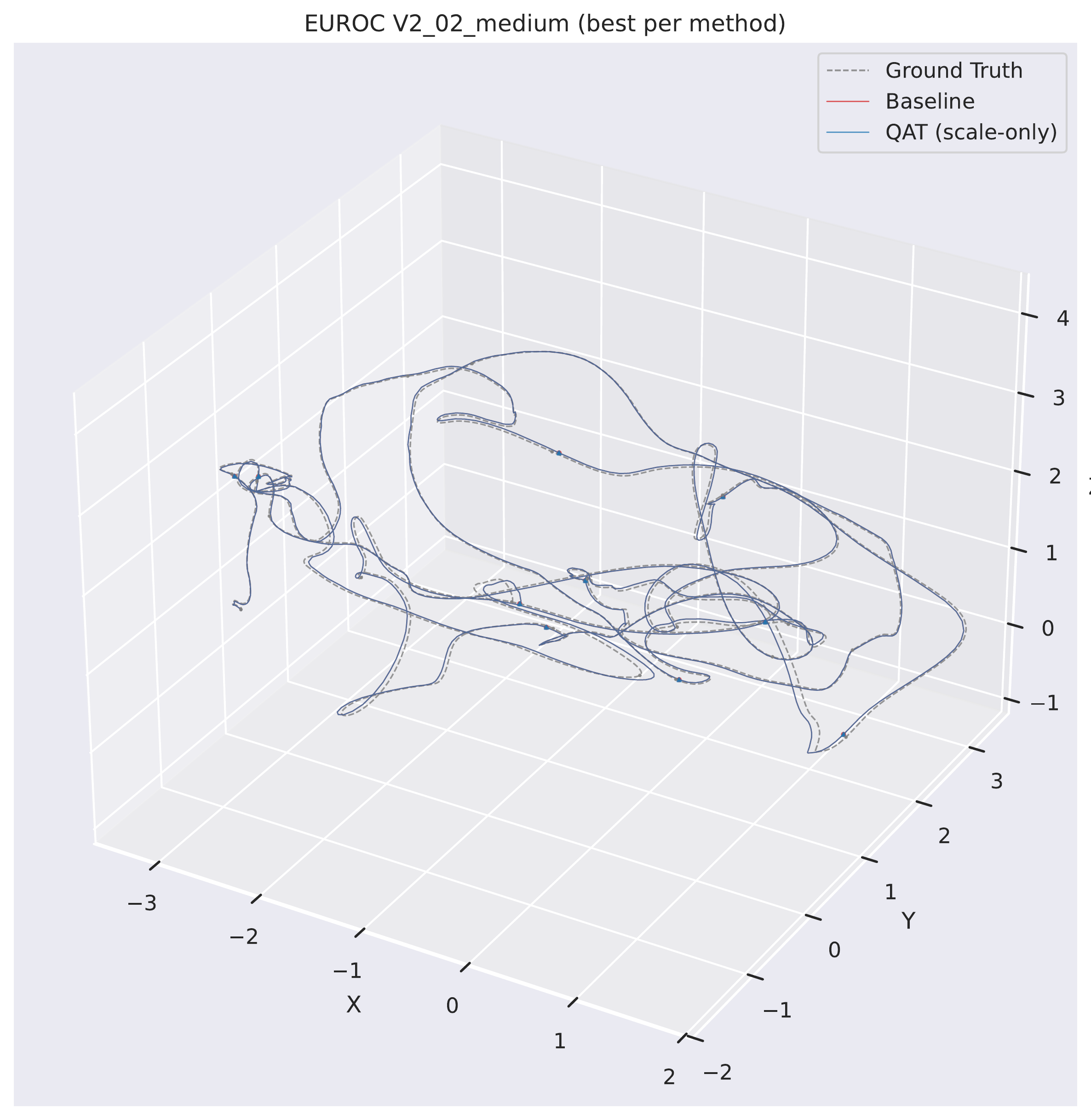}
  
  \vspace{1mm} 
  
  \includegraphics[width=\columnwidth]{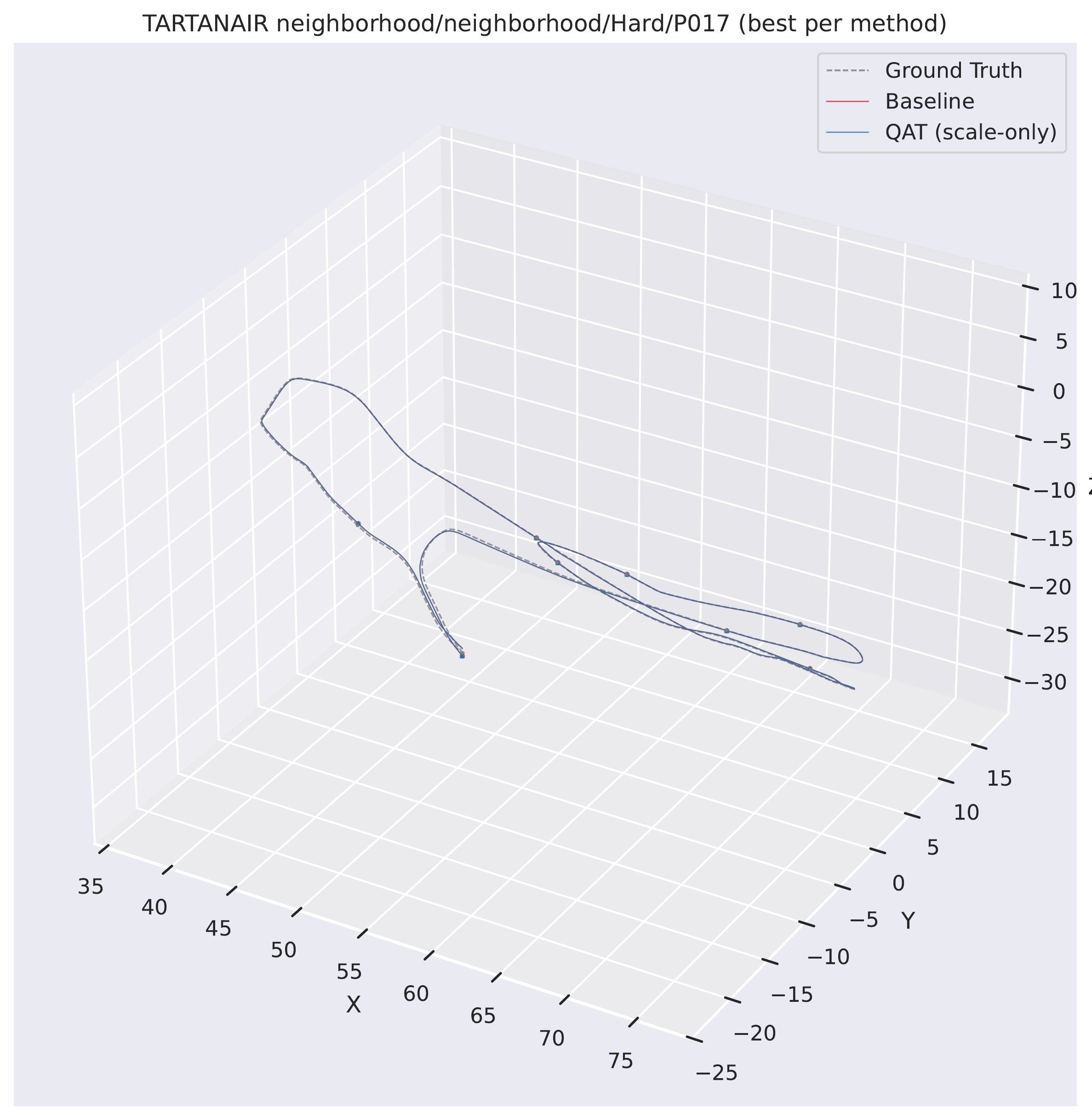}
  
  \caption{
    3D trajectory and map overlay comparisons on representative challenging sequences. 
    \textbf{Top}: EuRoC (\texttt{V2\_02\_medium}). 
    \textbf{Bottom}: TartanAir (\texttt{neighborhood\_Hard\_P017}). 
    The overlaid 3D point clouds demonstrate high geometric consistency.
  }
  \label{fig:combined_overlays}
\end{figure}

  \item \textbf{One-off initialization cost.} Due to CUDA context creation and kernel loading, TTFP can be slightly higher than the baseline, but this amortizes in long runs (see Fig.~\ref{fig:latency_bars} for latency comparison).
\end{itemize}


\subsubsection{Throughput and Latency}\label{subsec:throughput-latency}

The significant throughput and latency improvements summarized in our key findings are consistent across both datasets. This section delves into the specifics, supported by detailed per-sequence results (Table~\ref{tab:euroc_mh01} and Table~\ref{tab:tartanair_abf_hard}) and our timing methodology. Latency is collected by a unified \texttt{Timer} that prefers CUDA events and falls back to CPU clocks when needed:



\begin{table*}[!t]
  \centering
  \footnotesize
  \setlength{\tabcolsep}{7pt}
  \renewcommand{\arraystretch}{1.12}
  \begin{tabular}{@{}l r r r r r@{}}
    \toprule
    \textbf{Method} & \textbf{ATE (m, median)} & \textbf{FPS (avg)} & \textbf{P50 (ms)} & \textbf{P99 (ms)} & \textbf{Peak Reserved (GB)} \\
    \midrule
    Baseline & 0.0670 & 11.32 & 85.92 & 103.50 & 3.26 \\
    \textsc{QAT--DPVO} & 0.0827 & 11.15 & 86.95 & 105.20 & 1.96 \\
    \textsc{DPVO--QAT++} & 0.0583 & 14.92 & 65.20 & 76.20 & 2.03 \\
    \bottomrule
  \end{tabular}
  \caption{EuRoC (MH\_01\_easy): median ATE, average FPS, latency, and peak reserved memory.}
  \label{tab:euroc_mh01}
\end{table*}

\begin{table*}[!t]
  \centering
  \footnotesize
  \setlength{\tabcolsep}{7pt}
  \renewcommand{\arraystretch}{1.12}
  \begin{tabular}{@{}l r r r r r@{}}
    \toprule
    \textbf{Method} & \textbf{ATE (m, median)} & \textbf{FPS (avg)} & \textbf{P50 (ms)} & \textbf{P99 (ms)} & \textbf{Peak Reserved (GB)} \\
    \midrule
    Baseline & 0.0977 & 11.90 & 83.80 & 93.64 & 1.94 \\
    \textsc{QAT--DPVO} & 0.1378 & 11.77 & 84.65 & 94.84 & 2.91 \\
    \textsc{DPVO--QAT++} & 0.0305 & 15.98 & 63.16 & 70.87 & 1.02 \\
    \bottomrule
  \end{tabular}
  \caption{TartanAir (abandonedfactory/Hard/P011): median ATE, average FPS, latency, and peak reserved memory.}
  \label{tab:tartanair_abf_hard}
\end{table*}

\subsubsection{Memory Efficiency and Mechanism}\label{subsec:memory}

The substantial reduction in peak \texttt{Reserved} memory is attributed to the design of our fused CUDA path. Mechanistically, this path computes \texttt{softplus}+\texttt{clamp} for scales and performs Q/DQ within a single kernel on the GPU, reducing intermediate tensors and kernel launches, thus lowering allocator pressure and global-memory traffic. Convolutions always run on cuDNN float kernels, avoiding extra format/precision shuffles. In contrast, per-operator fake-quant triggers many small kernels and global memory round-trips, driving up the allocator's reserved pool. Distinguishing \texttt{Allocated} vs.\ \texttt{Reserved}: the former is the live tensor peak; the latter is the allocator's pool cap. Fusion reduces fragmentation and short-lived allocations (curbing \texttt{Reserved}), while mixed precision (FP16 features/caches, float convolutions/BA) also lowers \texttt{Allocated}.\supercite{TeedDeng2023DPVO}

\subsubsection{Trajectory Accuracy}\label{subsec:accuracy}

The high trajectory accuracy, which remains comparable to the baseline, validates the design of our ``scale-only'' QAT policy combined with a full-precision back-end. This approach ensures that front-end perturbations are controlled and absorbed without breaking global geometric consistency. A EuRoC APE computation excerpt:


\subsubsection{Layer-wise Sensitivity and Strategy}\label{subsec:sensitivity}

From scale statistics, early convolutions (e.g., input layers) and $1\!\times\!1$ channel-mapping layers have more conservative weight-scale distributions. Early layers act directly on pixel intensities; $1\!\times\!1$ layers aggregate across channels, where quantization error may be amplified or redistributed. A practical strategy is a semi-custom mixed precision: keep a few sensitive layers in float while using fused Q/DQ elsewhere, improving ATE robustness and tail latency with negligible throughput loss.

\subsection{Ablation Studies}\label{sec:ablations}

We evaluate the specific contribution of system-level fusion by running the same QAT-trained model through two inference paths:

\begin{itemize}[leftmargin=1.2em]
  \item \textbf{Per-operator fake-quant (\textsc{QAT--DPVO}).} Standard PyTorch path invoking separate Q and DQ operators around each layer; convolutions remain in float.
  \item \textbf{Fused CUDA front end (\textsc{DPVO--QAT++}).} Our path fuses \texttt{softplus(scale)+clamp} with Q/DQ into a single kernel; convolutions remain in float.
\end{itemize}

The core question: do frequent micro-kernel launches and global memory transfers in the standard path negate theoretical low-precision advantages; and can fusion reclaim them as measurable end-to-end gains?

\subsubsection{Path Contrast and Cost Model}\label{subsubsec:path-model}

Standard per-operator Q/DQ launches a separate kernel and global-memory
round-trip for each operation, which introduces notable system overhead,
especially on bandwidth-limited or high-latency workloads. By contrast,
our fused strategy consolidates multiple steps to shrink this
“non-convolution” overhead. The expected gains should appear as reduced
P99 tail latency and lower peak \texttt{Reserved} memory.



\subsubsection{EuRoC: Sequence Comparison}\label{subsubsec:euroc-abs}

The empirical results, detailed in Table~\ref{tab:euroc_abs} and Table~\ref{tab:euroc_delta}, strongly substantiate our cost model. The data confirms that our fused path (\textsc{DPVO--QAT++}) effectively absorbs per-layer overhead, converting theoretical advantages into significant performance gains: in terms of throughput (FPS), there is a stable improvement of approximately +3.0 to +3.6 FPS acroPer-ss all sequences; for tail latency (P99), the P99 decreases by 18.3 to 27.3 ms for each sequence; regarding memory reserved (Reserved), the peak Reserved memory is reduced from 3.26 GB to 2.03 GB, consistently decreasing by approximately 1.23 GB (≈ −37.7\%); most sequences perform on par with or better than the baseline.

Conversely, the per-operator path (\textsc{QAT--DPVO}) introduces additional tail latency ($\Delta$P99 of $+3$ to $+8$\,ms), validating our analysis. Accuracy deviations ($\Delta$ATE) are minor for most sequences, with the exception of \texttt{V2\_03\_difficult}, which we attribute to its specific sensitivities and believe could be improved with broader training coverage.

\begin{table*}[!t]
  \centering
  \scriptsize
  \setlength{\tabcolsep}{3.5pt}
  \renewcommand{\arraystretch}{1.1}
  \begin{tabular}{@{}l r r r r r r r r r r r r@{}}
    \toprule
    \multirow{2}{*}{\textbf{Seq.}} & \multicolumn{3}{c}{\textbf{FPS}} & \multicolumn{3}{c}{\textbf{P99 (ms)}} & \multicolumn{3}{c}{\textbf{Reserved (GB)}} & \multicolumn{3}{c}{\textbf{ATE (m)}} \\
    \cmidrule(lr){2-4}\cmidrule(lr){5-7}\cmidrule(lr){8-10}\cmidrule(lr){11-13}
     & Base & QAT & CUDA & Base & QAT & CUDA & Base & QAT & CUDA & Base & QAT & CUDA \\
    \midrule
    MH\_01\_easy & 11.32 & 11.15 & 14.92 & 103.50 & 105.20 & 76.20 & 3.26 & 1.96 & 2.03 & 0.0670 & 0.0827 & 0.0583 \\
    MH\_02\_easy & 11.34 & 11.13 & 14.98 & 100.50 & 106.18 & 75.40 & 3.26 & 1.96 & 2.03 & 0.0526 & 0.0586 & 0.0602 \\
    MH\_04\_difficult & 12.21 & 11.86 & 15.16 & 91.75 & 96.97 & 72.70 & 3.26 & 1.96 & 2.03 & 0.1344 & 0.1569 & 0.1683 \\
    MH\_05\_difficult & 11.90 & 11.50 & 15.23 & 93.66 & 99.77 & 74.00 & 3.26 & 1.96 & 2.03 & 0.0987 & 0.0928 & 0.1150 \\
    V1\_01\_easy & 12.21 & 11.71 & 15.46 & 90.42 & 97.07 & 71.90 & 3.26 & 1.96 & 2.03 & 0.0526 & 0.0601 & 0.0714 \\
    V1\_03\_difficult & 11.49 & 11.03 & 14.56 & 96.03 & 104.42 & 76.20 & 3.26 & 1.96 & 2.03 & 0.0940 & 0.1236 & 0.1096 \\
    V2\_01\_easy & 12.62 & 12.12 & 15.74 & 89.04 & 96.07 & 70.70 & 3.26 & 1.96 & 2.03 & 0.0498 & 0.0710 & 0.0440 \\
    V2\_03\_difficult & 11.33 & 11.03 & 14.35 & 99.96 & 102.99 & 76.90 & 3.26 & 1.96 & 2.03 & 0.2280 & 1.0347 & 0.7403 \\
    \bottomrule
  \end{tabular}
  \caption{EuRoC per-sequence (absolute): Baseline / QAT / CUDA.}
  \label{tab:euroc_abs}
\end{table*}

\begin{table*}[!t]
  \centering
  \scriptsize
  \setlength{\tabcolsep}{5pt}
  \renewcommand{\arraystretch}{1.1}
  \begin{tabular}{@{}l r r r r r@{}}
    \toprule
    \textbf{Seq.} & $\Delta$FPS (C--B) & $\Delta$P99 (C--B, ms) & $\Delta$Reserved (C--B, GB) & $\Delta$ATE (C--B, m) & $\Delta$P99 (Q--B, ms) \\
    \midrule
    MH\_01\_easy & +3.60 & -27.30 & -1.23 & -0.0087 & +1.70 \\
    MH\_02\_easy & +3.64 & -25.10 & -1.23 & +0.0076 & +5.68 \\
    MH\_04\_difficult & +2.95 & -19.05 & -1.23 & +0.0339 & +5.22 \\
    MH\_05\_difficult & +3.33 & -19.66 & -1.23 & +0.0163 & +6.11 \\
    V1\_01\_easy & +3.25 & -18.52 & -1.23 & +0.0188 & +6.65 \\
    V1\_03\_difficult & +3.07 & -19.83 & -1.23 & +0.0156 & +8.39 \\
    V2\_01\_easy & +3.12 & -18.34 & -1.23 & -0.0058 & +7.03 \\
    V2\_03\_difficult & +3.02 & -23.06 & -1.23 & +0.5123 & +3.03 \\
    \bottomrule
  \end{tabular}
  \caption{EuRoC per-sequence deltas: $\Delta$\,CUDA--Base and related.}
  \label{tab:euroc_delta}
\end{table*}

\subsubsection{TartanAir: Sequence Comparison}\label{subsubsec:tartanair-abs}

Considering space constraints, we report only representative results; the relevant results are summarized in Table~\ref{tab:tartanair_abs} (absolute metrics) and Table~\ref{tab:tartanair_delta} (delta metrics).
From Table~\ref{tab:tartanair_abs} and Table~\ref{tab:tartanair_delta}, \textsc{DPVO--QAT++} achieves stable frame-rate improvements of \(+3.82\text{--}+4.22\) FPS across all listed sequences; for tail latency (P99), each sequence shows a decrease of \(19.46\text{--}24.70\)~ms; for memory usage, the peak \texttt{Reserved} drops from \(1.94\)~GB to \(1.02\)~GB for every sequence (\(\Delta\text{Reserved}=-0.92\)~GB, \(\approx -47.4\%\)). In terms of accuracy (ATE), most sequences perform on par with or better than the baseline, while a few exhibit degradations. Compared with the per-operator path, \(\Delta\mathrm{P99}(\mathrm{Q{-}B})\) is positive for all sequences (\(+0.80\text{--}+3.53\)~ms), reaffirming that the per-operator approach not only fails to improve end-to-end latency but also introduces additional tail latency.

\begin{table*}[!t]
  \centering
  \scriptsize
  \setlength{\tabcolsep}{3.5pt}
  \renewcommand{\arraystretch}{1.1}
  \begin{tabular}{@{}l r r r r r r r r r r r r@{}}
    \toprule
    \multirow{2}{*}{\textbf{Seq.}} & \multicolumn{3}{c}{\textbf{FPS}} & \multicolumn{3}{c}{\textbf{P99 (ms)}} & \multicolumn{3}{c}{\textbf{Reserved (GB)}} & \multicolumn{3}{c}{\textbf{ATE (m)}} \\
    \cmidrule(lr){2-4}\cmidrule(lr){5-7}\cmidrule(lr){8-10}\cmidrule(lr){11-13}
     & Base & QAT & CUDA & Base & QAT & CUDA & Base & QAT & CUDA & Base & QAT & CUDA \\
    \midrule
    abandonedfactory/Easy/P011 & 11.77 & 11.63 & 15.59 & 96.22 & 97.17 & 72.92 & 1.94 & 2.91 & 1.02 & 0.2212 & 0.2421 & 0.1583 \\
    abandonedfactory/Hard/P011 & 11.90 & 11.77 & 15.98 & 93.64 & 94.84 & 70.87 & 1.94 & 2.91 & 1.02 & 0.0977 & 0.1378 & 0.0305 \\
    gascola/Easy/P008 & 12.02 & 11.70 & 16.08 & 93.83 & 97.36 & 72.02 & 1.94 & 2.91 & 1.02 & 0.1653 & 0.1181 & 0.2052 \\
    neighborhood/Hard/P017 & 12.00 & 11.67 & 16.22 & 94.74 & 97.74 & 72.03 & 1.94 & 2.91 & 1.02 & 0.1641 & 0.1461 & 0.1982 \\
    oldtown/Easy/P007 & 12.46 & 12.29 & 16.47 & 85.17 & 86.72 & 65.71 & 1.94 & 2.91 & 1.02 & 0.2279 & 0.2263 & 0.1939 \\
    soulcity/Easy/P012 & 11.85 & 11.74 & 16.04 & 91.62 & 92.58 & 71.94 & 1.94 & 2.91 & 1.02 & 0.1605 & 0.1394 & 0.1440 \\
    seasonsforest\_winter/Easy/P009 & 11.91 & 11.75 & 15.98 & 97.53 & 98.37 & 72.83 & 1.94 & 2.91 & 1.02 & 0.0948 & 0.1179 & 0.0603 \\
    office/Hard/P007 & 11.54 & 11.37 & 15.41 & 96.05 & 96.85 & 72.84 & 1.94 & 2.91 & 1.02 & 0.0195 & 0.0335 & 0.0657 \\
    \bottomrule
  \end{tabular}
  \caption{TartanAir representative subset (absolute): Baseline / QAT / CUDA.}
  \label{tab:tartanair_abs}
\end{table*}

\begin{table*}[!t]
  \centering
  \scriptsize
  \setlength{\tabcolsep}{6pt}
  \renewcommand{\arraystretch}{1.1}
  \begin{tabular}{@{}l r r r r r@{}}
    \toprule
    \textbf{Seq.} & $\Delta$FPS (C--B) & $\Delta$P99 (C--B, ms) & $\Delta$Reserved (C--B, GB) & $\Delta$ATE (C--B, m) & $\Delta$P99 (Q--B, ms) \\
    \midrule
    abandonedfactory/Easy/P011 & 3.82 & -23.30 & -0.92 & -0.0629 & 0.95 \\
    abandonedfactory/Hard/P011 & 4.08 & -22.77 & -0.92 & -0.0672 & 1.20 \\
    gascola/Easy/P008 & 4.06 & -21.81 & -0.92 & +0.0399 & 3.53 \\
    neighborhood/Hard/P017 & 4.22 & -22.71 & -0.92 & +0.0341 & 3.00 \\
    oldtown/Easy/P007 & 4.01 & -19.46 & -0.92 & -0.0340 & 1.55 \\
    soulcity/Easy/P012 & 4.19 & -19.68 & -0.92 & -0.0165 & 0.96 \\
    seasonsforest\_winter/Easy/P009 & 4.07 & -24.70 & -0.92 & -0.0345 & 0.84 \\
    office/Hard/P007 & 3.87 & -23.21 & -0.92 & +0.0462 & 0.80 \\
    \bottomrule
  \end{tabular}
  \caption{TartanAir representative subset (deltas): $\Delta$\,CUDA--Base and related.}
  \label{tab:tartanair_delta}
\end{table*}

\subsubsection{Component and Numeric-Path Ablation}\label{subsubsec:components}

We clarify the components and numeric paths:
\begin{itemize}[leftmargin=1.2em]
  \item \textbf{Quantization coverage.} All \texttt{Conv2d} layers in the front-end \textit{Patchifier}, including the initial $7{\times}7$, $3{\times}3$ in residual blocks, $1{\times}1$ downsampling, and tail $1{\times}1$ convolutions (22 layers total).
  \item \textbf{Operators and kernels.} In the fused path, one CUDA kernel handles scale transform (\texttt{softplus+clamp}) and fake Q/DQ. Convolutions are float and executed by cuDNN. For FP16 inputs we use a safe FP16$\rightarrow$FP32$\rightarrow$FP16 fallback to improve numeric stability.
  \item \textbf{Quantization policy.} Symmetric INT8 (S8, $q_{\max}{=}127$); per-channel weights and per-tensor activations. Backprop uses a straight-through estimator (STE). Scales are computed in FP32, with stricter lower bounds on the FP16 path to avoid underflow/saturation.
\end{itemize}

\section{Conclusion and Future Work}\label{sec:conclusion}

\subsection{Conclusion}\label{sec:conclusion-summary}
This paper addresses the deployment challenge of deep learning-based vSLAM on resource-constrained platforms by proposing \textsc{DPVO--QAT++}, an integrated optimization framework that combines heterogeneous quantization, Quantization-Aware Training (QAT), and native GPU acceleration. Building upon the state-of-the-art DPVO, we systematically overhaul it from both algorithmic and system-level perspectives, validating the following core conclusions:

\begin{enumerate}
  \item \textbf{Heterogeneous precision is key.} Applying pseudo-quantization (symmetric INT8 with learned scales) to the computationally intensive front-end network while keeping the geometric optimization back-end (Bundle Adjustment, BA) in full precision (\texttt{FP32}) is crucial for striking an effective balance between efficiency and accuracy. At the code level, the front-end operates under \texttt{autocast} (primarily \texttt{FP16}) and is explicitly cast back to \texttt{FP32} before the BA step to ensure numerical stability for the solver.

  \item \textbf{A geometry-oriented QAT strategy is effective.} Our \textit{scale-only} learning approach, combined with a teacher-student distillation strategy that aligns Patchifier features using MSE and cosine similarity losses, effectively compensates for quantization errors. This preserves the geometric representation capabilities of the pre-trained model, ultimately achieving empirically equivalent Absolute Trajectory Error (ATE).

  \item \textbf{Algorithm-system co-design is necessary.} Without a custom CUDA implementation, per-operator pseudo-quantization typically fails to deliver end-to-end acceleration. Our method completes the quantization process using a few custom GPU kernels (one for computing scales, and others for quantizing activations and weights, respectively) before invoking ATen/cuDNN for convolution. Compared to the per-operator path, this significantly reduces the overhead from small kernel launches and memory transfers, successfully converting theoretical speedups into measured performance gains (FPS$\uparrow$, P99$\downarrow$, Memory$\downarrow$).
\end{enumerate}

\textbf{The significance of this work} lies in bridging the critical "deployment gap" between deep learning SLAM models and the practical constraints of real-world robotic systems. We demonstrate that, through a hardware-software co-optimization design methodology, it is possible to achieve the real-time performance and resource efficiency required for on-board processing while maintaining the high accuracy of complex models like DPVO. This research provides a validated and reproducible paradigm for deploying other complex perception algorithms that follow a "deep front-end + classic back-end" architecture, advancing the development of autonomous navigation technology for deployment on embedded platforms.

\subsection{Limitations and Future Work}\label{sec:future-work}
Despite promising results, this work has several limitations. \textbf{First}, our current implementation relies on floating-point pseudo-quantization to simulate low-precision behavior, and the performance impact of true integer-only arithmetic has not yet been evaluated on target hardware. \textbf{Second}, all experiments were conducted on a PC-grade GPU, and the performance gains require further validation on target embedded platforms, such as the NVIDIA Jetson series. \textbf{Finally}, the scope of our work is confined to the visual odometry (VO) front-end and does not yet constitute a full SLAM system with global optimization capabilities such as loop closure and relocalization.

These limitations also directly point to our future work directions:
\begin{itemize}[leftmargin=1.2em]
  \item \textbf{Deeper Quantization Exploration.} Building on the success of the floating-point pseudo-quantization front-end, we will systematically evaluate the impact of true low-precision integer arithmetic (\texttt{INT8}/\texttt{INT4}) on the trajectory accuracy and stability of VO/SLAM systems.\supercite{Peng2024QVIO} This includes developing comprehensive kernel coverage, weight layout optimization, and reordering strategies.

  \item \textbf{Deployment on Target-Specific Hardware.} We plan to deeply integrate the current framework with inference engines like NVIDIA TensorRT, aiming to migrate our design philosophy of "fused low-overhead quantization, pre-allocation, zero-copy, and pipelining" to generate optimized computation graphs for embedded hardware like Jetson, thereby further unlocking bandwidth and parallelism.

  \item \textbf{Extension to Full SLAM.} The current work focuses on the VO front-end. The next step is to integrate it into a full SLAM system with loop closure and a global pose graph optimization. We will evaluate the impact of quantization on global consistency and relocalization capabilities and explore heterogeneous precision strategies for the loop closure module.

  \item \textbf{Method Generalization.} The "heterogeneous quantization with hardware-software co-design" principle proposed in this paper is broadly applicable. It can be transferred to other robotics perception tasks that follow a "deep front-end + classic back-end" paradigm to achieve robust end-to-end system gains.
  
  \item \textbf{Building an Automated Co-Design Toolchain.} A key future direction is to encapsulate the proposed hardware-software co-design methodology into an automated optimization and deployment toolchain. This toolchain aims to take a user's pre-trained floating-point model as input and, through an automated process, complete three core tasks: (1) generate the corresponding QAT training code; (2) execute the quantization-aware training and optimization; and (3) finally, compile the optimized model into a high-efficiency inference engine composed of fused CUDA kernels and INT8 operators. This would thereby shield developers from the complexities of low-level quantization and hardware optimization, providing a "one-click" path from model to high-performance edge deployment.
\end{itemize}

\flushend


\bibliographystyle{IEEEtran}
\bibliography{references}

\end{document}